\documentclass[sigconf]{acmart}
\AtBeginDocument{%
  \providecommand\BibTeX{{%
    \normalfont B\kern-0.5em{\scshape i\kern-0.25em b}\kern-0.8em\TeX}}}

\copyrightyear{2026}
\acmYear{2026}
\setcopyright{cc}
\setcctype{by}
\acmConference[KDD '26]{Proceedings of the 32nd ACM SIGKDD Conference on Knowledge Discovery and Data Mining V.2}{August 09--13, 2026}{Jeju Island, Republic of Korea}
\acmBooktitle{Proceedings of the 32nd ACM SIGKDD Conference on Knowledge Discovery and Data Mining V.2 (KDD '26), August 09--13, 2026, Jeju Island, Republic of Korea}
\acmDOI{10.1145/3770855.3818152}
\acmISBN{979-8-4007-2259-2/2026/08}

\usepackage{balance}
\usepackage{graphics}
\usepackage{graphicx}
\usepackage{mdwlist}

\usepackage{amsmath,amsfonts,bm}
\usepackage{pifont}
\DeclareMathOperator*{\argmax}{arg\,max}
\newcommand{\cmark}{\ding{51}} % check mark
\newcommand{\xmark}{\ding{55}} % cross mark (optional)
\usepackage{IEEEtrantools}

\usepackage{amsthm}
\usepackage{siunitx}

\usepackage{caption}
\usepackage{subcaption}

\usepackage{array}
\usepackage{makecell}
\usepackage{multirow}
\usepackage{multicol}
\usepackage{url}
\usepackage{enumitem}

\begin{document}

\newcommand{\goal}[1]{ {\noindent {$\Rightarrow$} \em {#1} } }

% ****************** TITLE ****************************************

\title[PathCRF: Ball-Free Soccer Event Detection via Possession Path Inference from Player Trajectories]{PathCRF: Ball-Free Soccer Event Detection via \\ Possession Path Inference from Player Trajectories}

%\numberofauthors{6}

\author{Hyunsung Kim}
\affiliation{
  \institution{KAIST}
  \city{Daejeon}
  \country{Republic of Korea}
}
\additionalaffiliation{%
  \institution{Fitogether Inc.}
  \city{Seoul}
  \country{Republic of Korea}
}
\email{hyunsung.kim@kaist.ac.kr}

\author{Kunhee Lee}
\affiliation{
    \institution{KAIST}
    \city{Daejeon}
    \country{Republic of Korea}
}
\email{kunhee8@kaist.ac.kr}

\author{Sangwoo Seo}
\affiliation{
    \institution{KAIST}
    \city{Daejeon}
    \country{Republic of Korea}
}
\email{sangwooseo@kaist.ac.kr}

\author{Sang-Ki Ko}
\affiliation{
    \institution{University of Seoul}
    \city{Seoul}
    \country{Republic of Korea}
}
\email{sangkiko@uos.ac.kr}

\author{Jinsung Yoon}
\affiliation{
    \institution{Fitogether Inc.}
    \city{Seoul}
    \country{Republic of Korea}
}
\email{jinsung.yoon@fitogether.com}

\author{Chanyoung Park}
\affiliation{
    \institution{KAIST}
    \city{Daejeon}
    \country{Republic of Korea}
}
\email{cy.park@kaist.ac.kr}

\renewcommand{\shortauthors}{Hyunsung Kim et al.}
    
\begin{abstract}
	Despite recent advances in AI, event data collection in soccer still relies heavily on labor-intensive manual annotation. Although prior work has explored automatic event detection using player and ball trajectories, ball tracking also remains difficult to scale due to high infrastructural and operational costs. As a result, comprehensive data collection in soccer is largely confined to top-tier competitions, limiting the broader adoption of data-driven analysis in this domain. To address this challenge, this paper proposes \textbf{PathCRF}, a framework for detecting on-ball soccer events using only player tracking data. We model player trajectories as a fully connected dynamic graph and formulate event detection as the problem of selecting exactly one edge corresponding to the current possession state at each time step. To ensure logical consistency of the resulting edge sequence, we employ a Conditional Random Field (CRF) that forbids impossible transitions between consecutive edges, where emission and transition scores are dynamically computed from edge embeddings produced by a socio-temporal backbone architecture. During inference, the most probable edge sequence is obtained via Viterbi decoding, and events such as ball controls or passes are detected whenever the selected edge changes between adjacent time steps. Experiments show that PathCRF produces accurate, logically consistent possession paths, enabling reliable downstream analyses while substantially reducing the need for manual event annotation.
The source code is available at \url{https://github.com/hyunsungkim-ds/pathcrf.git}.
\end{abstract}

\begin{CCSXML}
<ccs2012>
   <concept>
       <concept_id>10002951.10003227.10003236</concept_id>
       <concept_desc>Information systems~Spatial-temporal systems</concept_desc>
       <concept_significance>500</concept_significance>
       </concept>
   <concept>
       <concept_id>10010147.10010178.10010187.10010190</concept_id>
       <concept_desc>Computing methodologies~Probabilistic reasoning</concept_desc>
       <concept_significance>300</concept_significance>
       </concept>
   <concept>
       <concept_id>10010147.10010257.10010293.10010294</concept_id>
       <concept_desc>Computing methodologies~Neural networks</concept_desc>
       <concept_significance>300</concept_significance>
       </concept>
 </ccs2012>
\end{CCSXML}

\ccsdesc[500]{Information systems~Spatial-temporal systems}
\ccsdesc[300]{Computing methodologies~Probabilistic reasoning}
\ccsdesc[300]{Computing methodologies~Neural networks}

\keywords{Sports Analytics; Event Detection; Multi-Agent Trajectory Modeling; Sequence Labeling; Conditional Random Fields}

%% page.s

\maketitle

\section{Introduction}
\label{sec:intro}
The massive data collection in professional soccer has fundamentally transformed how decisions are made in this domain, shifting analysis from subjective judgments toward data-driven approaches. Contemporary soccer analytics primarily relies on two types of data~\cite{AnzerABBBD25}. The first is event data, which records on-ball actions such as passes, dribbles, and shots. The second is tracking data, which provides high-frequency spatiotemporal coordinates of all players and the ball throughout a match. Since they capture complementary aspects of the game, comprehensive match analysis generally requires access to both data modalities.

However, acquiring both data remains costly and challenging. While recent advances in wearable and vision-based systems have made player tracking relatively inexpensive and automated, ball tracking remains substantially more expensive and difficult to scale. Due to the ball’s small size, high speed, and frequent occlusions, vision-based ball tracking often suffers from missed detections or false positives, requiring dense multi-camera setups and labor-intensive manual corrections \cite{MaksaiWF16,RenOJX09,WangASF14,WuXLMP20}. Sensor-based solutions \cite{FIFA22,Monnit22} further raise deployment costs, as they require meticulous calibration and expensive stadium-wide infrastructure.

In parallel, even in the era of advanced AI, event data collection still relies heavily on manual annotation. Human annotators must watch entire matches and record every event's timestamp, location, involved players, and category. Several studies~\cite{BischofbergerBS24,LinkH17,VidalCodinaEEB22} have attempted to automatically detect events from player and ball trajectories, but they still depend on accurate ball tracking, limiting their scalability. Consequently, pipelines for collecting complete sets of event and tracking data remain largely confined to top-tier competitions, and have not scaled to lower or youth divisions.

To overcome these limitations, prior work has explored inferring ball-related information directly from player trajectories. Kim et al.~\cite{KimCKYK23} and Capellera et al.~\cite{CapelleraFRAM24,CapelleraRFA25} infer ball positions or classify ball states using only player trajectories. However, these approaches do not explicitly enforce physical constraints, often producing implausible ball trajectories or unrealistically frequent possession changes. Other studies~\cite{PeralCRFMA25,PeralCFRA26} incorporate visual cues such as player image crops extracted from video to improve accuracy, but they use tactical camera footage that always captures the entire pitch, which is substantially harder to obtain than standard broadcasting video or tracking data. This requirement runs counter to the goal of democratizing automated data collection across lower-tier competitions.

In this paper, we propose \textbf{PathCRF}, a framework for detecting on-ball soccer events solely from player trajectories. Rather than estimating ball positions or identifying ball states, we formulate event detection as the problem of inferring a \emph{possession path}, indicating how the ball moves between players over time. To this end, player trajectories are represented as a fully connected dynamic graph, and at each time step the model selects exactly one edge corresponding to the current possession state: a self-loop for dribbling, or a directed sender–receiver edge for a kick in flight.

To ensure that the inferred possession path is logically consistent, we construct a Conditional Random Field (CRF) \cite{LaffertyMP01} on top of a neural backbone. The backbone jointly encodes the dynamic interactions between players over time via socio-temporal attention \cite{LeeLKKCT19} and produces a sequence of candidate edge embeddings. The CRF dynamically computes emission and transition scores from these edge embeddings, where logically impossible edge-to-edge transitions are masked with a large negative score so that any sequence containing illegal transitions is assigned a very low likelihood. At inference time, the most probable sequence is obtained via Viterbi decoding~\cite{Viterbi67}, and on-ball events are detected whenever the selected edge changes between consecutive time steps.

Experimental results show that PathCRF identifies the correct possession edge among $26^2=676$ candidate edges with an accuracy of 69.64\%, and achieves F1-score of 75.69\% in event detection.
% Notably, the high recall implies that human annotators only need to supplement the remaining 22\% of events, which substantially reduces the manual effort required for event data collection.
Furthermore, we show that a wide range of downstream metrics, including event heatmaps, team possession statistics, and pass networks, are closely approximated using the detected events, facilitating reliable match analysis without manual event annotation.

In summary, this paper proposes a framework for detecting on-ball soccer events using only player tracking data, without requiring expensive ball-tracking hardware or labor-intensive annotation. The main contributions are as follows:
\begin{itemize}[leftmargin=0.5cm]
    \item Formulating soccer event detection as a sequential edge selection problem over a dynamic graph.
    \item Introducing neural CRF-based sequence inference, used primarily in natural language processing (NLP), into sports analytics to enforce domain-specific physical constraints.
    \item Demonstrating that key event-based analytics, including spatial tendencies, team possession statistics, and passing patterns, can be reliably approximated without manual event annotation.
    \item Releasing the source code using publicly available tracking data~\cite{BassekRWM25}, so that anyone can easily reproduce the results.
\end{itemize}

% \section{Related Work}
% \label{sec:survey}
% \input{contents/020survey}

\section{Proposed Framework}
\label{sec:framework}
\begin{figure*}[t!]
    \centering
    \vspace{-0.5em}
    \includegraphics[width=0.98\textwidth]{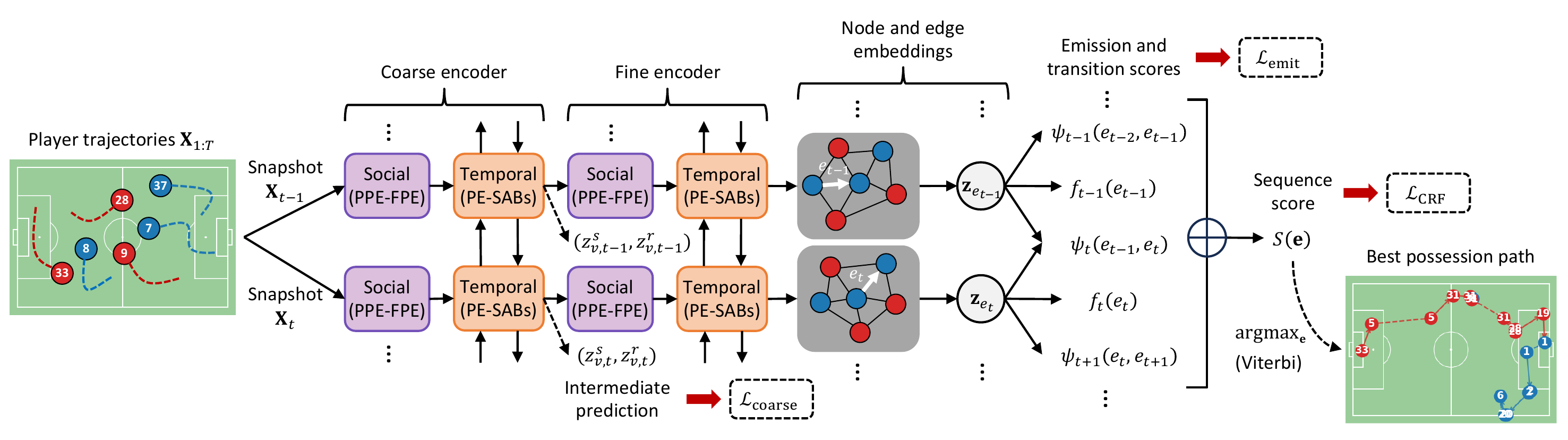}
    \vspace{-1em}
    \caption{Overall architecture of PathCRF.}
    \vspace{-0.5em}
    \label{fig:overview}
\end{figure*}

In this section, we formulate soccer event detection as a path inference problem over a fully connected dynamic graph (Section \ref{sec:problem}) and present a framework consisting of a neural backbone (Section \ref{sec:backbone}), a Dynamic Masked Conditional Random Field (CRF) (Section \ref{sec:crf}), and an event detection mechanism (Section \ref{sec:event}). The overall architecture is illustrated in Fig.~\ref{fig:overview}.

\subsection{Problem Formulation}
\label{sec:problem}
Our ultimate objective is to detect on-ball soccer events using only player tracking data. To this end, we model player trajectories as a fully connected dynamic graph and formulate event detection as the problem of inferring a \emph{possession path}, represented as a sequence of edges that describe the ball possession states over time.

Formally, given a set of player trajectories $\mathbf{X}_{1:T}=\{\mathbf{x}_{v,1:T}\}_{v \in \mathcal{V}}$, we represent each snapshot as a fully connected directed graph over the extended node set $\widetilde{\mathcal{V}}$. This set is defined as the union of the player set $\mathcal{V}$ and the set $\mathcal{O}$ of four \emph{outside} nodes, each corresponding to the ball leaving the pitch through one of the four boundary lines (left, right, top, and bottom)\footnote{Since $|\mathcal{V}|=22$ in standard soccer matches without red cards, the total number of nodes is $|\widetilde{\mathcal{V}}|=26$ in most cases.}. Then, the ball possession state at each time step $t$ corresponds to exactly one edge $e_t=(u_t,v_t)$ between nodes in $\widetilde{\mathcal{V}}$, where a self-loop $(u_t,u_t)$ indicates that the player $u_t$ is in control of the ball, while a directed edge $(u_t,v_t)$ with $u_t \neq v_t$ represents the ball traveling from sender $u_t$ to receiver $v_t$.

The goal of the model is therefore to select the edge sequence $\hat{\mathbf{e}} =(\hat{e}_1, \ldots, \hat{e}_T)$ indicating the ball possession path from player trajectories alone. Once the possession path is obtained, on-ball events are detected by identifying time steps where the selected edge changes, i.e., $\hat{e}_{t-1} \neq \hat{e}_t$. Each detected event is then categorized according to the type of the newly selected edge, yielding \emph{control}, \emph{kick}, or \emph{out-of-play} events (see Section~\ref{sec:event} for details).

\subsection{Backbone Architecture}
\label{sec:backbone}
To extract rich socio-temporal representations of player interactions, we adopt an encoder design inspired by ROLAND \cite{YouDL22} that combines social module to capture multi-agent interactions and a temporal module to encode sequential dependencies. Following TranSPORTmer \cite{CapelleraFRAM24}, we stack two of these encoders to progressively refine the game contexts from coarse to fine levels. The resulting node embeddings are subsequently transformed into edge embeddings, which serve as the fundamental features for scoring edge sequences in the CRF described in Section~\ref{sec:crf}.

\subsubsection{Socio-temporal encoder blocks.}
\label{sec:encoder}
Let $\mathbf{H}^{(l-1)}_t \in \mathbb{R}^{|\widetilde{\mathcal{V}}| \times d}$ denote the input node features for the $l$-th encoder at time $t$ ($l=1,2$), where $\mathbf{H}^{(0)}_t = \mathbf{X}_t = \{\mathbf{x}_{v,t}\}_{v\in\widetilde{\mathcal{V}}}$. Each encoder consists of a social module followed by a temporal module, jointly capturing spatial interactions among players and their evolution over time.

\vspace{1mm}
\noindent \textbf{Social module.}
Following Kim et al. \cite{KimCKYK23}, we model multi-agent interactions at each time step using a combination of a partially permutation-equivariant (PPE) network and a fully permutation-equivariant (FPE) network. The PPE encoder processes nodes within each group (i.e., either team or the set of outside nodes) separately using an Induced Set Attention Block (ISAB) \cite{LeeLKKCT19} to capture intra-group dynamics, respecting the fact that players within a team are semantically unordered yet distinct from those in the opposing team. Complementarily, the FPE module processes all nodes through a single ISAB to encode global inter-group contexts. The outputs of these PPE and FPE networks are concatenated to form node representations for the snapshot at $t$:
\begin{equation}
    \widetilde{\mathbf{H}}^{(l)}_t
    = \left\{ \tilde{\mathbf{h}}^{(l)}_{v,t} \right\}_{v\in\widetilde{\mathcal{V}}}
    = \left[ \text{PPE}\left(\mathbf{H}^{(l-1)}_t\right); \text{FPE}\left(\mathbf{H}^{(l-1)}_t\right) \right] \in \mathbb{R}^{|\widetilde{\mathcal{V}}| \times d}.
    \label{eq:social}
\end{equation}

\vspace{1mm}
\noindent \textbf{Temporal module.}
The temporal module aggregates node representations across time to capture the evolution of interactions:
\begin{equation}
    \mathbf{h}^{(l)}_{v,1:T}
    = \text{Temporal}\left(\tilde{\mathbf{h}}^{(l)}_{v,1:T} \right) \in \mathbb{R}^{T \times d},
    \label{eq:temporal}
\end{equation}
which yields the node embeddings $\mathbf{H}^{(l)}_t = \{\mathbf{h}^{(l)}_{v,t}\}_{v\in\widetilde{\mathcal{V}}}$ that serve as the input for the next stage. In principle, this module can be instantiated with any sequence model, but in this work, we adopt the architecture proposed in TranSPORTmer \cite{CapelleraFRAM24} that combines positional encoding (PE) \cite{VaswaniSPUJGKP17} with stacked Set Attention Blocks (SABs) \cite{LeeLKKCT19} as a concrete instantiation.

Empirically, in Section \ref{sec:ablation_backbone}, we observe that this attention-based temporal module achieves performance comparable to its Bi-LSTM \cite{HochreiterS97} counterpart. We attribute this behavior to the nature of our task: soccer possession dynamics are largely driven by short-term continuity, where the most recent context is typically the most informative. As a result, both recurrent models with an explicit recency bias and attention-based models that learn which time steps to focus on tend to yield similar performance in this setting.

\subsubsection{Intermediate supervision.}
\label{sec:intermed}
To encourage the backbone to learn possession-related semantics early in the network, we introduce an auxiliary supervision task using intermediate node embeddings $\mathbf{H}^{(1)}_{t}$ produced by the first (coarse) encoder, inspired by prior work \cite{ZhanZYSL19,KimCKYK23}. Specifically, each node embedding $\mathbf{h}^{(1)}_{v,t} \in \mathbb{R}^d$ is projected into a two-dimensional vector comprising a sender logit and a receiver logit via a Multilayer Perceptron (MLP):
\begin{equation}
    (z^s_{v,t}, z^r_{v,t}) = \text{MLP}_{\text{coarse}} \left(\mathbf{h}^{(1)}_{v,t}\right) \in \mathbb{R}^2.
    \label{eq:mlp_coarse}
\end{equation}
We then apply the softmax normalization across all nodes for the sender and receiver logits, respectively, to obtain the probabilities of each node being the sender or the receiver of the ball at time $t$. The cross-entropy loss between these probabilities and the ground-truth sender/receiver labels is incorporated as an auxiliary term into the final training objective detailed in Section \ref{sec:training}.

\subsubsection{Edge embedding construction.}
\label{sec:edge_embed}
From the second (fine) encoder, we obtain refined node embeddings $\mathbf{H}^{(2)}_{t} = \{\mathbf{h}^{(2)}_{v,t}\}_{v \in \widetilde{\mathcal{V}}}$. To transition from node-level representations to candidate possession states, we construct a $d'$-dimensional embedding for each directed edge $e=(u,v)$ by concatenating the embeddings of its source (sender) and destination (receiver) nodes, followed by an MLP:
\begin{equation}
    \mathbf{z}_{e,t} = \text{MLP}_{\text{edge}} \left(\left[\mathbf{h}^{(2)}_{u,t}; \mathbf{h}^{(2)}_{v,t}\right]\right) \in \mathbb{R}^{d'}.
    \label{eq:mlp_edge}
\end{equation}
These edge embeddings serve as the core features used to compute the CRF scores described in Section~\ref{sec:scoring}.

\subsection{Dynamic Masked Conditional Random Field}
\label{sec:crf}
\begin{figure}[!bt]
\centering
\subfloat[Ground-truth possession path and events\label{fig:true_events}]{
    \hspace*{-1em}
	\includegraphics[width=.5\textwidth]{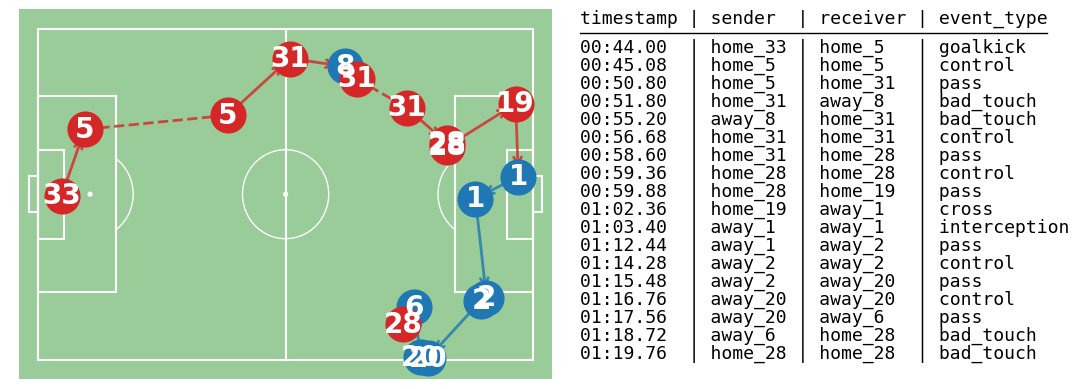}
}
\vspace{0.5em}
\subfloat[Independent edge prediction per time step without a CRF \label{fig:indep_events}]{
    \hspace*{-1em}
	\includegraphics[width=.49\textwidth]{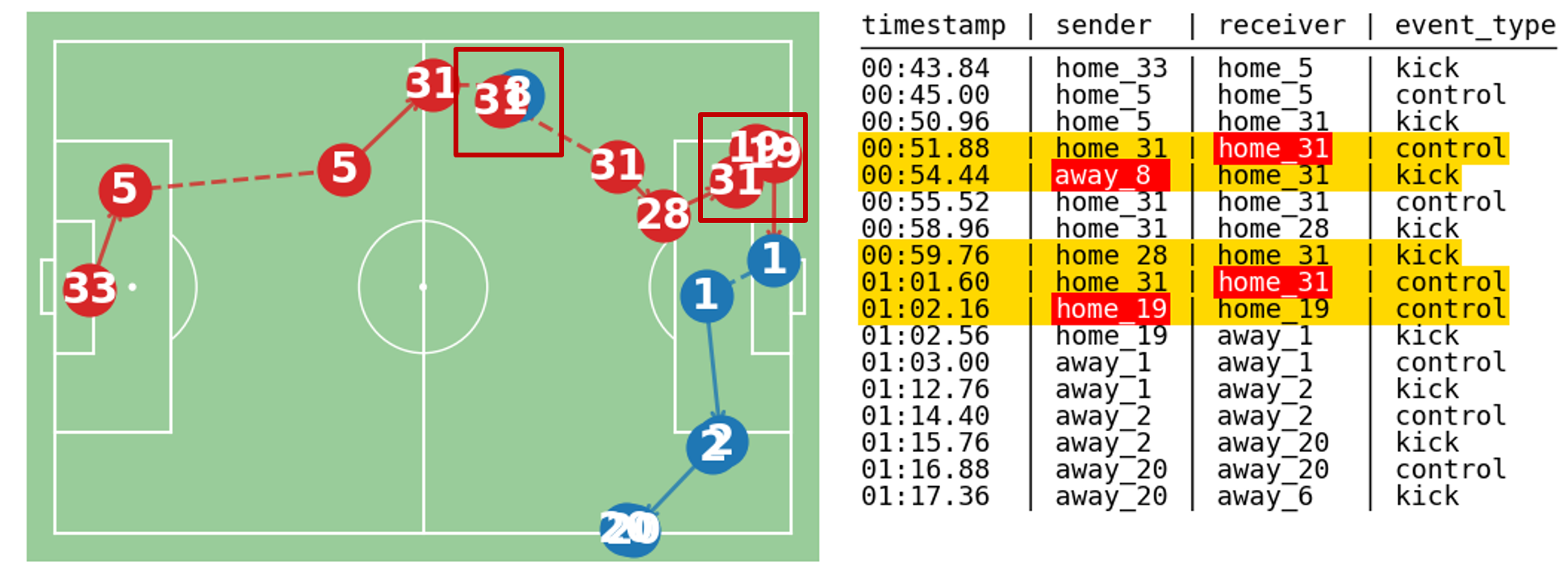}
}
\vspace{0.5em}
\subfloat[Prediction by the proposed PathCRF \label{fig:crf_events}]{
    \hspace*{-1em}
	\includegraphics[width=.49\textwidth]{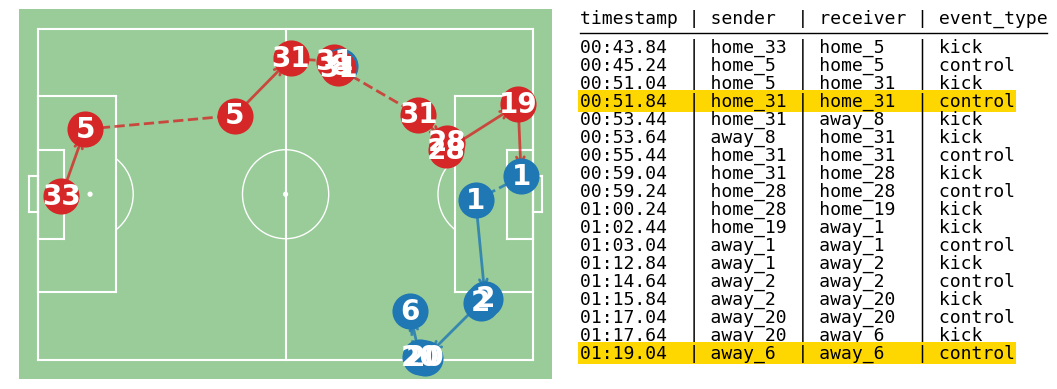}
}
\vspace{-0.5em}
\caption{Comparison of the ground-truth possession path and events with model predictions on an in-play segment of the test dataset. Predicted events with incorrect sender or receiver attributes are highlighted in yellow, and physically impossible transitions are highlighted in red.}
\vspace{-0.5em}
\label{fig:events}
\end{figure}

Given the edge embeddings $\mathbf{z}_{e,t}$ produced by the backbone, a straightforward approach to infer a possession path is to compute edge logits independently at each time step and form a sequence by selecting the highest-scoring edge via an argmax operation. However, such per-time-step classification ignores semantic dependencies between consecutive possession states, often producing logically inconsistent edge sequences. For instance, a model might select a self-loop for player $u$ at time $t$ and immediately predict another edge $(v,w)$ where $u \neq v$ at time $t+1$ without any intermediate pass edge $(u,v)$, implying a physically impossible ``teleportation'' of the ball (see the red-highlighted players in Fig.~\ref{fig:indep_events}). To resolve this, we employ a Conditional Random Field (CRF)~\cite{LaffertyMP01} to strictly enforce domain-specific transition constraints.

% For example, after a dribbling state represented by a self-loop of player $u$, only edges originating from $u$ are allowed. Similarly, after a pass state represented by a directed edge $(u,v)$, the next state should either remain the same edge $(u,v)$ or transition to an edge originating from the receiver $v$.

\subsubsection{Scoring edge sequences.}
\label{sec:scoring}
Unlike the independent classification approach that assigns a probability to each edge per time step, the CRF framework models the probability of the entire edge sequence $\mathbf{e}=(e_1, \ldots, e_T)$ given the input observations $\mathbf{X}=\mathbf{X}_{1:T}$. The probability of a sequence $\mathbf{e}$ is defined as
\begin{equation}
    p(\mathbf{e}|\mathbf{X}) = \frac{\exp(S(\mathbf{e}))}{\sum_{\mathbf{e'}}\exp(S(\mathbf{e'}))}
\end{equation}
where $S(\mathbf{e})$ is the score of the sequence, computed as the sum of emission and transition scores over all time steps:
\begin{equation}
    S(\mathbf{e}) = f_1(e_1) + \sum_{t=2}^T \left( f_t(e_t) + \psi_t(e_{t-1},e_t) \right).
    \label{eq:score}
\end{equation}

The emission score $f_t(e_t)$ indicates the local likelihood of edge $e_t$ being the current possession state. We compute this score by projecting the edge embedding $\mathbf{z}_{e,t}$ obtained from Eq.~\ref{eq:mlp_edge}:
\begin{equation}
    f_t(e) = \text{MLP}_{\text{emit}}(\mathbf{z}_{e,t}) \in \mathbb{R}.
    \label{eq:mlp_emit}
\end{equation}

The transition score $\psi_t(e_{t-1},e_t)$ measures the possibility of transitioning from the previous edge $e_{t-1}$ to the current edge $e_t$. Standard CRFs typically parameterize this score using a static transition matrix $\Psi$, where each entry $\psi_{ij}$ is a learnable weight representing the likelihood of transition from state $i$ to $j$~\cite{HuangXY15,MaH16}.

However, this static parametrization is ill-suited for capturing the time-varying nature of game dynamics. For instance, consider a long pass corresponding to the directed edge $(u,v)$. In the early phase of the pass, the probability of the identity transition $(u,v) \rightarrow (u,v)$ should be high as the ball is likely to be still traveling toward the receiver. As time progresses and the ball approaches the receiver $v$, however, the likelihood of transitions into a reception state such as $(v,v)$ and $(v,w)$ should increase. A static transition matrix cannot adapt to such evolving spatiotemporal context.

% Second, it violates permutation-equivariance, which is a fundamental requirement for multi-agent trajectory modeling. Since edges are indexed by sender and receiver identities, the row and column positions of states in $\Psi$ depends entirely on the arbitrary ordering of players in the input. As a result, transition scores become tied to index assignments rather than semantic roles, making the model fail to learn meaningful and consistent transition patterns.

To address this limitation, we compute transition scores \emph{dynamically} from edge embeddings produced by the backbone. Specifically, for a given transition from edge $e'$ at $t-1$ to edge $e$ at $t$, we concatenate their embeddings and pass them through a shared MLP:
\begin{equation}
    \psi_t(e',e) = \text{MLP}_{\text{trans}}([\mathbf{z}_{e',t-1};\mathbf{z}_{e,t}]) \in \mathbb{R}.
    \label{eq:mlp_trans}
\end{equation}
This allows the transition scores to adapt to the evolving game context at each time step, thereby improving the event detection performance as discussed in Section \ref{sec:results}.

\subsubsection{Masking impossible transitions.}
\label{sec:masking}
Another key feature of our approach is the imposition of hard constraints on these transition scores. 
%\textcolor{blue}{(hereafter, \emph{transition constraints})}. 
First, we define the \emph{set of allowed transitions} $\mathcal{A}$ as the union of the following subsets:
\begin{itemize}[leftmargin=0.5cm]
    \item $\mathcal{A}^{\text{id}} = \{(u,v) \rightarrow (u,v): u,v \in \widetilde{\mathcal{V}}\}$,
    representing the continuation of the current state,
    \item $\mathcal{A}^{\text{kick}} = \{(u,u) \rightarrow (u,v): u \in \mathcal{V}, v \in \widetilde{\mathcal{V}}, u \neq v\}$,
    representing a player $u$ releasing the ball to pass to $v$,
    \item $\mathcal{A}^{\text{rec}} = \{(u,v) \rightarrow (v,w): u,v \in \mathcal{V}, w \in \widetilde{\mathcal{V}}, u \neq v \}$,
    representing the receiver $v$ either controlling the ball ($w=v$) or making a one-touch kick ($w \neq v$), and
    \item $\mathcal{A}^{\text{out}} = \{(u,o) \rightarrow (o,o): u \in \mathcal{V}, o \in \mathcal{O}\}$, representing the ball going out of play\footnote{Since we perform training and inference only on windows inside in-play segments as described in Section \ref{sec:data}, the self-loop $(o,o)$ of each outside node $o \in \mathcal{O}$ serves as an absorbing state of our edge-to-edge transition system.}.
\end{itemize}
Then, following the strategy of Masked CRF~\cite{WeiQHS21}, we mask all transitions except for those in $\mathcal{A}$ by a fixed, large negative score (e.g., $-10^4$). This mechanism effectively prunes illegal paths from the search space and forces the model to produce only logically consistent sequences.

\subsubsection{Model training.}
\label{sec:training}
We train the entire framework end-to-end by minimizing the negative log-likelihood of the ground-truth edge sequence $\mathbf{e}^* = (e^*_1, \ldots, e^*_T)$ given the input trajectories $\mathbf{X}$:
\begin{equation}
\mathcal{L}_{\text{CRF}} = -\log p(\mathbf{e}^* | \mathbf{X}) = \log Z - S(\mathbf{e}^*)
\label{eq:crf_loss}
\end{equation}
where $S(\mathbf{e}^*)$ is the score of the ground-truth sequence computed by Eq.~\ref{eq:score} and $Z = \sum_{\mathbf{e'}} \exp(S(\mathbf{e'}))$ is the partition function that normalizes the probability distribution. Following the standard formulation of CRFs~\cite{LaffertyMP01}, we efficiently compute $\log Z$ using the forward algorithm based on dynamic programming, where the detailed procedure is elaborated on Appendix \ref{sec:forward}.

To further stabilize training and encourage discriminative feature learning, we add two auxiliary loss terms to the training objective: the \emph{intermediate classification loss} $\mathcal{L}_{\text{coarse}} = \mathcal{L}_{\text{coarse}}^s + \mathcal{L}_{\text{coarse}}^r$ described in Section \ref{sec:intermed} and the \emph{emission classification loss} $\mathcal{L}_{\text{emit}}$, which computes the cross-entropy between the emission logits $f_t(e)$ and the true possession state $e^*_t$. The final objective is constructed as a weighted sum of the three loss terms:
\begin{equation}
    \mathcal{L} = \mathcal{L}_{\text{CRF}} + \lambda_1 \mathcal{L}_{\text{coarse}} + \lambda_2 \mathcal{L}_{\text{emit}},
    \label{eq:loss}
\end{equation}
where $\lambda_1$ and $\lambda_2$ are hyperparameters balancing the contributions of the auxiliary terms.

\subsubsection{Viterbi decoding.}
\label{sec:decoding}
During the inference phase, the CRF framework identifies the single most likely edge sequence $\hat{\mathbf{e}} = \arg\max_{\mathbf{e}} S(\mathbf{e})$ using the Viterbi algorithm~\cite{LaffertyMP01,Viterbi67}. By incorporating the hard constraints described in Section \ref{sec:masking}, the decoding process automatically prunes impossible transitions from the search space, ensuring that the inferred path is logically consistent with all transitions belonging to $\mathcal{A}$.  For further details, see Appendix \ref{sec:viterbi}.

\subsection{Event Detection}
\label{sec:event}
Once the best edge sequence $\hat{\mathbf{e}} = (\hat{e}_1, \ldots, \hat{e}_T)$ is obtained via Viterbi decoding, we deterministically extract discrete soccer events by identifying time steps where the possession state changes (i.e., $\hat{e}_{t-1} \neq \hat{e}_{t}$). At each change-point $t$, the detected event is categorized based on the topology of the initiated edge $\hat{e}_{t}$ as follows:
\begin{itemize}[leftmargin=0.5cm]
    \item \textbf{Control:} A ball control is detected at time $t$ when the state transitions to a player self-loop $(u,u)$ with $u \in \mathcal{V}$ at $t$.
    \item \textbf{Kick:} A kick is detected at time $t$ when a state changes to an edge $(u,v)$ with $u \in \mathcal{V}, v \in \widetilde{\mathcal{V}}, u \neq v$ at $t$. Note that this event can occur immediately after a ball control by $u$ (controlled kick) or a kick towards $u$ (one-touch kick).
    \item \textbf{Out-of-play:} An out-of-play event is detected when the state transitions to an outside node self-loop $(o,o)$ with $o \in \mathcal{O}$ at $t$, indicating the ball has crossed the pitch boundary.
\end{itemize}

Fig.~\ref{fig:crf_events} visualizes a representative example of the inferred possession path and the corresponding events detected by our framework. In contrast to the non-CRF baseline in Fig.~\ref{fig:indep_events}, which produces illegal transitions where the receiver of the previous event does not match the sender of the next event (red-highlighted case), PathCRF explicitly enforces transition constraints that guarantee a globally consistent possession sequence. Moreover, our dynamic estimation of emission and transition scores not only eliminates such impossible event patterns but also improves prediction accuracy, yielding fewer mislabeled events as indicated by the reduced proportion of yellow-highlighted errors compared to Fig.~\ref{fig:indep_events}. Overall, PathCRF effectively segments continuous game flow into on-ball events solely from player trajectories.

\section{Experiments}
\label{sec:experiments}
In this section, we evaluate PathCRF through extensive experiments on publicly available soccer tracking data \cite{BassekRWM25}. Section~\ref{sec:data} describes the dataset and the preprocessing pipeline. Section~\ref{sec:baselines} and Section~\ref{sec:metrics} introduce baseline methods and evaluation metrics, respectively. Section~\ref{sec:results} reports the main experimental results, and Section~\ref{sec:ablation_backbone} provides an ablation study on backbone design choices. Lastly, Section~\ref{sec:runtime} analyzes the runtime efficiency of the framework.

\subsection{Data Preparation}
\label{sec:data}
As professional soccer clubs and data providers typically keep tracking data proprietary, complete sets of event and tracking data are rarely publicly available. To ensure full reproducibility, we conduct experiments exclusively on the Sportec Open DFL Dataset~\cite{BassekRWM25}, which provides high-quality event and tracking data from seven matches of German Bundesliga 1 and 2. Despite the limited scale of the data, PathCRF achieves strong performance even when trained on only five matches, as reported in Section \ref{sec:results}. We further verify its robustness and generalizability on an additional non-public dataset in Appendix~\ref{app:robustness}.

% which is one of the very few publicly available datasets offering full observability of all players in the pitch at high spatiotemporal accuracy. This dataset provides event and tracking data from seven matches of German Bundesliga 1 and 2, and we split them into 5:1:1 for training, validation, and testing.

As the manual annotations often contain imprecise timestamps, we synchronize event timing based on player and ball trajectories using ELASTIC \cite{KimCSBYP25}. In addition, since the raw event data miss some ball touches, we detect abrupt directional changes in the ball trajectory using the Ramer–Douglas–Peucker (RDP) algorithm \cite{DouglasP73,Ramer72} as candidate ball-touch points. Then, we align them with the synchronized events via the Needleman-Wunsch algorithm~\cite{NeedlemanW70}, and insert unmatched points as additional ball-touch events. Finally, from this refined set of events, we generate ground-truth edge labels indicating the possession state per time step.

We restrict training and inference to in-play segments, since no on-ball events occur during stoppages. Following prior work~\cite{KimCKYK23}, we define each continuous in-play segment as an \emph{episode}, and apply 10-second sliding windows with a stride of 5 frames in each episode when constructing the training dataset. The tracking data are downsampled from 25\,FPS to 5\,FPS for computational efficiency, resulting in 50 time steps per window, and predictions are later upsampled to 25\,FPS for downstream analyses. The performance-efficiency trade-off of this downsampling is analyzed in Section~\ref{sec:runtime}. Dataset statistics are summarized in Table~\ref{tab:data}.

\subsection{Baseline Methods}
\label{sec:baselines}
We compare the proposed framework against a range of baselines to evaluate the impact of structured inference, dynamic transition scoring, and explicit masking of illegal transitions. All baselines share the same backbone architecture described in Section \ref{sec:backbone}.

\begin{itemize}[leftmargin=0.5cm]
    \item \textbf{Ball Trajectory Postprocessing (Ball TP):} A two-stage pipeline that first regresses the 2D ball trajectory from player tracking data, and then detect ball possession using the rule-based heuristics proposed in Ball Radar \cite{KimCKYK23}.
    \item \textbf{Non-CRF:} The backbone is trained only with cross-entropy losses (i.e., $\mathcal{L}_{\text{coarse}}$ and $\mathcal{L}_{\text{emit}}$ in Eq. \ref{eq:loss}), and the possession path is obtained by independently selecting the edge with the highest logit at each time step.
    \item \textbf{Greedy Constrained Decoding (GCD):} Uses the same Non-CRF model as above, but greedily enforces transition constraints at inference time. The first time step selects the argmax edge in terms of output logit, and each subsequent step selects the argmax edge only among those reachable from the previously selected edge under the transition rules defined in Section \ref{sec:masking}.
    \item \textbf{Viterbi Constrained Decoding (VCD):} Uses the same Non-CRF model as above, but applies Viterbi decoding \cite{Viterbi67}, obtaining the globally optimal sequence while satisfying the predefined transition constraints.
    \item \textbf{Static Dense CRF:} A standard CRF with a static, dense transition matrix learned independently of the input, without masking illegal transitions \cite{HuangXY15}.
    \item \textbf{Static Masked CRF:} A static CRF that explicitly masks illegal transitions \cite{WeiQHS21} based on Section~\ref{sec:masking}.
    \item \textbf{Dynamic Dense CRF:} Dynamically computes transition scores as in Eq.~\ref{eq:mlp_trans}, but does not mask illegal transitions.
    \item \textbf{Dynamic Masked CRF (ours):} Combines dynamic transition scoring with explicit masking of illegal transitions.
\end{itemize}

\begin{table}[t!]
\centering
\caption{Statistics of the dataset splits.}
\vspace{-0.5em}
\small
\begin{tabular}{c|ccccc}
\toprule
\textbf{Split}  & \textbf{Matches} & \textbf{Episodes} & \textbf{Events} & \textbf{Frames} & \textbf{Windows} \\
\midrule
\textbf{Train.} & 5 & 379   & 8,792 & 364,003 & 54,683  \\
\textbf{Valid.} & 1 & 71    & 1,902 & 80,076 & 12,620  \\
\textbf{Test}   & 1 & 91    & 1,831 & 85,304 & ---     \\
\bottomrule
\end{tabular}
\vspace{-0.5em}
\label{tab:data}
\end{table}

\begin{table*}[t]
\centering
\caption{Performance comparison of baseline methods on the test data.}
\vspace{-0.7em}
\small
\setlength{\tabcolsep}{3pt}
\begin{tabular}{l|cccc|ccc}
\toprule
\textbf{Structuring method} 
& \makecell[c]{\textbf{Edge} \\ \textbf{accuracy}}
& \makecell[c]{\textbf{Sender} \\ \textbf{accuracy}}
& \makecell[c]{\textbf{Receiver} \\ \textbf{accuracy}}
& \makecell[c]{\textbf{Violation} \\ \textbf{rate}} 
& \makecell[c]{\textbf{Event precision}}
& \makecell[c]{\textbf{Event recall}} 
& \textbf{Event F1} \\
\midrule
Ball TP
& 47.01\% & 58.58\% & 67.01\% & \textbf{0.00\%}
& 44.64\% (967/2166) & 52.81\% (967/1831) & 48.39\% \\

Non-CRF
& 69.24\% & 79.78\% & 81.62\% & 2.80\% 
& 58.77\% (1347/2292) & 73.57\% (1347/1831) & 65.34\% \\

Non-CRF \& GCD
& 66.34\% & 76.89\% & 78.19\% & \textbf{0.00\%} 
& 68.81\% (1339/1946) & 73.13\% (1339/1831) & 70.90\% \\

Non-CRF \& VCD 
& \textbf{69.68\%} & 80.16\% & \underline{81.97\%} & \textbf{0.00\%} 
& \underline{69.50\%} (1413/2033) & 77.17\% (1413/1831) & 73.14\% \\

Static Dense CRF 
& 69.41\% & \textbf{80.59\% } & 81.61\% & 2.18\% 
& 62.54\% (1399/2237) & 76.41\% (1399/1831) & 68.78\% \\

Static Masked CRF 
& 69.54\% & 80.29\% & 82.02\% & \textbf{0.00\%} 
& 68.57\% (1464/2135) & \textbf{79.96\%} (1464/1831)
& \underline{73.83\%} \\

Dynamic Dense CRF 
& 69.08\% & 79.56\% & 81.95\% & 2.28\% 
& 61.75\% (1377/2230) & 75.20\% (1377/1831) & 67.82\% \\

Dynamic Masked CRF 
& \underline{69.64\%} & \underline{80.44\%} & \textbf{82.28\%} & \textbf{0.00\%} 
& \textbf{73.18\%} (1435/1961) & \underline{78.37\%} (1435/1831) 
& \textbf{75.69\%} \\
\bottomrule
\end{tabular}
\label{tab:results}
\end{table*}

\subsection{Evaluation Metrics}
\label{sec:metrics}
We evaluate the proposed framework at both the edge level and the event level to assess the quality of the inferred possession paths and the resulting on-ball events.

\vspace{1mm}
\noindent \textbf{Edge-level metrics.} 
At the edge level, we evaluate the predicted possession states at each time step by reporting \emph{sender accuracy}, \emph{receiver accuracy}, and \emph{edge accuracy}. In addition, we measure the \emph{violation rate}, defined as the percentage of illegal transitions in the predicted edge sequence.

\vspace{1mm}
\noindent \textbf{Event-level metrics.} At the event level, we evaluate event detection performance using \emph{precision}, \emph{recall}, and \emph{F1-score}. To compute these metrics, we align the detected and ground-truth event sequences using the Needleman–Wunsch algorithm \cite{NeedlemanW70}, and count a detected event as correct if it is matched with a true event having (i) the same event type, (ii) the same acting player, and (iii) a temporal difference within one second.
Since the ground-truth events have fine-grained SPADL~\cite{DecroosBVD19} event types, we map them into the three simplified categories defined in Section \ref{sec:event} for consistent matching.
% Since the ground-truth events follow the SPADL schema~\cite{DecroosBVD19} with fine-grained event types, we map them into the three simplified categories defined in Section \ref{sec:event} for consistent matching.
% Specifically, we first explicitly insert an \emph{out-of-play} event after each event going out of the pitch. Then, we categorize in-play events followed by the same player as \emph{controls}, and those followed by a different player as \emph{kicks}.

\subsection{Main Results}
\label{sec:results}
Table \ref{tab:results} compares baseline methods at both the edge and event levels, where the best and runner-up results are highlighted in \textbf{bold} and \underline{underline}, respectively. Since all methods share the same backbone, edge-level accuracies are largely similar except for Ball TP. However, event-level performance varies significantly, underscoring the importance of structured inference for reliable event detection.

First, Ball TP performs markedly worse than edge-based approaches. Although it predicts ball trajectories with a localization error of \SI{2.74}{m}, it assigns possession states only through heuristic postprocessing (e.g., using ball acceleration and player-to-ball distance). This leads to a notable drop in event detection quality, which motivates our edge inference formulation.

The Non-CRF baseline, which selects edges independently per time step, produces 2.80\% illegal transitions and over-segments the sequence, leading to many spurious event detections and low precision (58.77\%). In contrast, simply applying constrained decoding at inference time entirely eliminates violations and improves event precision by more than 10\%p (i.e., up to 68.81\% for GCD and 69.50\% for VCD), demonstrating that enforcing transition constraints is critical for reliable event detection. Moreover, VCD slightly outperforms GCD, indicating the advantage of globally optimizing the entire sequence under constraints than greedy local decisions.

Compared to constrained decoding, Masked CRFs further improve performance by incorporating constraints during training. While constrained decoding has become more prevalent in recent NLP due to the huge cost of fine-tuning LLMs and the complexity of linguistic constraints \cite{HokampL17,SunLWHLD19,LuWWJKKBQYZSC22,GengJP023}, our setting differs in that models can be efficiently trained from scratch and the governing constraints are simple (as explicitly defined in Section \ref{sec:masking}), making CRF-based training both feasible and advantageous.

Lastly, the comparison between CRF variants highlights the necessity of explicit violation masking and dynamic transition scoring. Specifically, Dense CRFs without masking still produce illegal transitions and perform much worse in event detection than masked variants, indicating that CRF-based inference alone is insufficient without enforcing hard constraints. Furthermore, Dynamic Masked CRF achieves a much higher event precision (73.18\%) than Static Masked CRF (68.57\%). This is because static transition matrices always assign non-negligible probabilities to unnecessary state changes, resulting in far more frequent event detections (2,135) than the ground truth (1,831). In contrast, Dynamic Masked CRF adapts the likelihood of maintaining the current state to the context, yielding a more realistic number of events (1,961) and thereby improving the F1-score.

\begin{table*}[t]
\centering
\caption{Backbone design ablation study on the test data.}
\vspace{-0.7em}
\small
\setlength{\tabcolsep}{2pt}
\begin{tabular}{ccc|cccc|ccc}
\toprule
\makecell[c]{\textbf{Social} \\ \textbf{module}}
& \makecell[c]{\textbf{Temporal} \\ \textbf{module}}
& \makecell[c]{\textbf{Structuring} \\ \textbf{method}}
& \makecell[c]{\textbf{Edge} \\ \textbf{accuracy}}
& \makecell[c]{\textbf{Sender} \\ \textbf{accuracy}}
& \makecell[c]{\textbf{Receiver} \\ \textbf{accuracy}}
& \makecell[c]{\textbf{Violation} \\ \textbf{rate}} 
& \makecell[c]{\textbf{Event precision}}
& \makecell[c]{\textbf{Event recall}} 
& \textbf{Event F1} \\
\midrule

\multirow{3}{*}{\makecell[c]{SAB \\ (TranSPORTmer)}}
& \multirow{3}{*}{\makecell[c]{PE-SABs \\ (TranSPORTmer)}}
& Non-CRF
& 58.49\% & 70.98\% & 73.84\% & 4.45\%
& 46.14\% (1107/2399) & 60.46\% (1107/1831) & 52.34\% \\

& & Non-CRF \& VCD
& 59.08\% & 71.71\% & 74.29\% & \textbf{0.00\%}
& 59.41\% (1237/2082) & 67.56\% (1237/1831) & 63.23\% \\

& & Dynamic MCRF
& 62.97\% & 74.91\% & 77.98\% & \textbf{0.00\%}
& 66.51\% (1281/1926) & 69.96\% (1281/1831) & 68.19\% \\
\midrule

\multirow{3}{*}{\makecell[c]{PPE-FPE \\ (Ball Radar)}}
& \multirow{3}{*}{\makecell[c]{Bi-LSTM \\ (Ball Radar)}}
& Non-CRF
& 69.15\% & 80.01\% & 81.01\% & 2.17\%
& 61.50\% (1337/2174) & 73.02\% (1337/1831) & 66.77\% \\

& & Non-CRF \& VCD
& 69.32\% & 80.22\% & 81.15\% & \textbf{0.00\%}
& 68.89\% (1382/2006) & 75.48\% (1382/1831) & 72.04\% \\

& & Dynamic MCRF
& \textbf{70.39\%} & \textbf{80.98\%} & \textbf{82.30\%} & \textbf{0.00\%}
& \underline{72.67\%} (1428/1965) & \underline{77.99\%} (1428/1831) & \underline{75.24\%} \\
\midrule

\multirow{3}{*}{\makecell[c]{PPE-FPE \\ (Ball Radar)}}
& \multirow{3}{*}{\makecell[c]{PE-SABs \\ (TranSPORTmer)}}
& Non-CRF
& 69.24\% & 79.78\% & 81.62\% & 2.80\%
& 58.77\% (1347/2292) & 73.57\% (1347/1831) & 65.34\% \\

& & Non-CRF \& VCD
& \underline{69.68\%} & 80.16\% & 81.97\% & \textbf{0.00\%}
& 69.50\% (1413/2033) & 77.17\% (1413/1831) & 73.14\% \\

& & Dynamic MCRF
& 69.64\% & \underline{80.44\%} & \underline{82.28\%} & \textbf{0.00\%}
& \textbf{73.18\%} (1435/1961) & \textbf{78.37\%} (1435/1831) & \textbf{75.69\%} \\
\bottomrule
\end{tabular}
\label{tab:ablation_backbone}
\end{table*}

\subsection{Ablation Study on Backbone Design}
\label{sec:ablation_backbone}
To better understand how different backbone designs affect the performance, we conduct an ablation study by combining components proposed in two recent sports trajectory modeling frameworks: Ball Radar \cite{KimCKYK23} and TranSPORTmer \cite{CapelleraFRAM24}. Specifically, Ball Radar employs a social module combining PPE and FPE networks based on ISABs \cite{LeeLKKCT19} described in Section \ref{sec:encoder} with a temporal module using Bi-LSTM \cite{HochreiterS97}. Meanwhile, TranSPORTmer uses a vanilla SAB \cite{LeeLKKCT19} social module and an attention-based temporal module consisting of positional encoding (PE) \cite{VaswaniSPUJGKP17} and stacked SABs.

Based on these components, Table~\ref{tab:ablation_backbone} compares three variants:
\begin{itemize}[leftmargin=0.5cm]
    \item The TranSPORTmer backbone (SAB \& PE-SABs)
    \item The Ball Radar backbone (PPE-FPE \& Bi-LSTM)
    \item Our hybrid backbone, which adopts the Ball Radar's social module (PPE-FPE) but replaces the temporal module with TranSPORTmer's encoder (PE-SABs). As described in Section \ref{sec:encoder}, this configuration is used as the main backbone architecture throughout the paper.
\end{itemize}
The results show that replacing the TranSPORTmer's SAB-based social module with Ball Radar’s PPE-FPE consistently yields a substantial improvement, indicating that explicitly separating intra-team interactions (PPE) and global cross-team contexts (FPE) provides a stronger inductive bias than treating all agents uniformly with a single SAB. In contrast, once PPE-FPE is adopted, the choice of temporal module between Bi-LSTM and PE-SABs has only a marginal impact, where Bi-LSTM slightly improves edge-level accuracies and PE-SABs achieves slightly better event-level performance. This suggests that soccer possession dynamics are largely governed by short-term continuity, where recent context is often the most informative, making both recurrent and attention-based temporal modules similarly effective in this task.

\subsection{Runtime Analysis}
\label{sec:runtime}
To verify the computational efficiency of PathCRF for practical use, we measure the training and inference speed of its variants on a single NVIDIA GeForce RTX 4090 GPU (24\,GB). We also compare our default 5\,FPS downsampling setting against the full 25\,FPS version. To examine the performance-efficiency trade-off, Table~\ref{tab:runtime} reports the event F1 alongside the training and inference time.

\begin{table}[b!]
\centering
\small
\caption{Training and inference speed measured on a single NVIDIA GeForce RTX 4090 GPU (24\,GB). Inference metrics (i.e., episode and match latency) are reported as the mean $\pm$ standard deviation over 3 runs.}
\vspace{-0.5em}
\label{tab:runtime}
\setlength{\tabcolsep}{1.5pt}
\begin{tabular}{l|cccc}
\toprule
\textbf{Method (FPS)} & \makecell{\textbf{Event}\\\textbf{F1}} & \makecell{\textbf{Training time}\\\textbf{(s/epoch)}} & \makecell{\textbf{Episode}\\\textbf{latency (s)}} & \makecell{\textbf{Match}\\\textbf{latency (s)}} \\
\midrule
Non-CRF (5)  & 65.34\% & $184.20 \pm 0.34$    & $0.2155 \pm 0.0013$ & $34.57 \pm 0.06$ \\
Static MCRF (5)  & 73.83\% & $241.70 \pm 0.47$    & $0.2104 \pm 0.0005$ & $20.17 \pm 0.05$ \\
Dynamic MCRF (5)  & 75.69\% & $565.41 \pm 4.19$    & $0.2399 \pm 0.0009$ & $22.88 \pm 0.08$ \\
Dynamic MCRF (25) & 77.21\% & $6726.95 \pm 27.80$  & $0.5272 \pm 0.0028$ & $49.28 \pm 0.27$ \\
\bottomrule
\end{tabular}
\end{table}

As shown in the table, the CRF-based variants require longer training time than the Non-CRF baseline, as they involve an additional CRF layer and the forward algorithm described in Appendix~\ref{sec:forward}. However, their inference speed remains comparable, resulting in all variants processing a full 90-minute match in under one minute. This is far shorter than the match duration itself, confirming that inference is not a practical bottleneck.

We also evaluate PathCRF applied to 25\,FPS tracking data without downsampling. While the setting improves event F1 from 75.69\% to 77.21\%, the training time per epoch increases by more than ten times (565.41 to 6726.95 seconds), and this gap would grow further with larger training data. Given the marginal accuracy gain relative to the substantial computational cost, we adopt 5\,FPS as the default configuration throughout the paper.

\section{Practical Applications}
\label{sec:applications}
In this section, we showcase a wide range of practical applications of the proposed PathCRF. First, we demonstrate that PathCRF can closely approximate diverse downstream analyses without requiring any manual event annotation, including spatial event heatmaps (Section \ref{sec:heatmaps}), team possession statistics (Section \ref{sec:poss_stats}), and pass networks (Section \ref{sec:passmap}). Furthermore, when completely accurate event data are still required for fine-grained scene analysis, Section \ref{sec:semi_annot} discusses how our framework can substantially reduce human workload by supporting semi-automated event data collection.

\begin{figure}[tbh]
  \centering
  \begin{minipage}[t]{0.45\textwidth}
    \includegraphics[width=\linewidth]{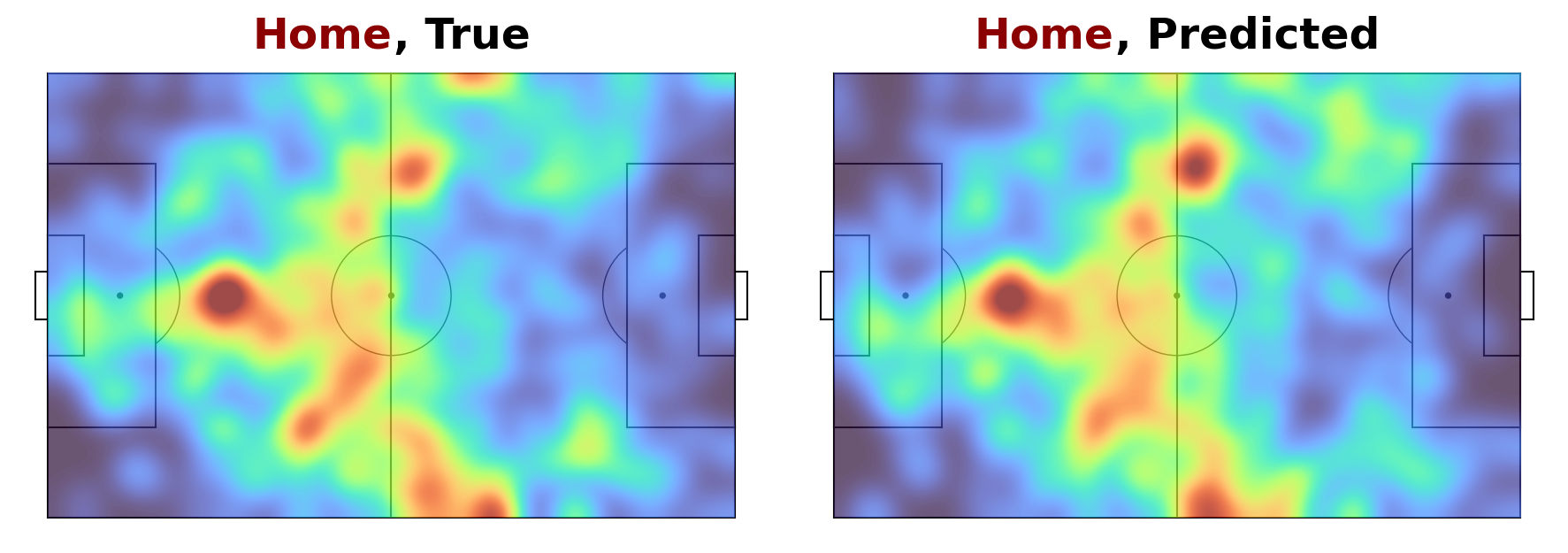}
  \end{minipage}
  \hfill
  \begin{minipage}[t]{0.45\textwidth}
    \includegraphics[width=\linewidth]{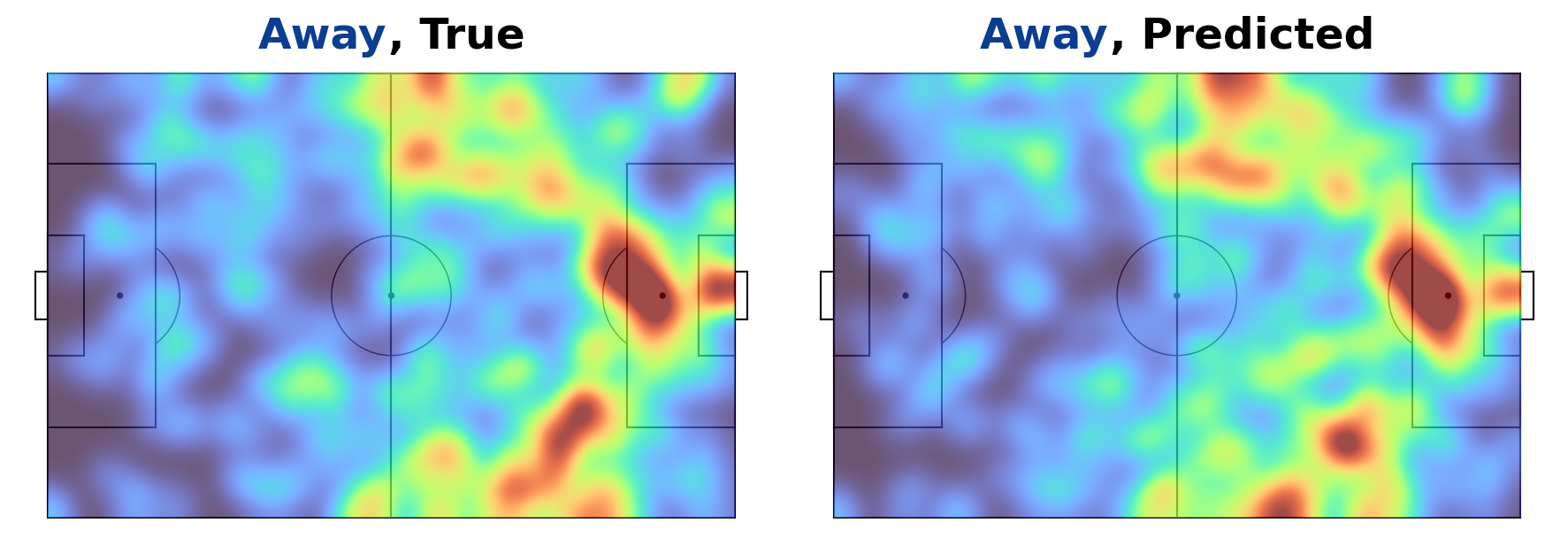}
  \end{minipage}
  \hfill
  \begin{minipage}[t]{0.45\textwidth}
    \includegraphics[width=\linewidth]{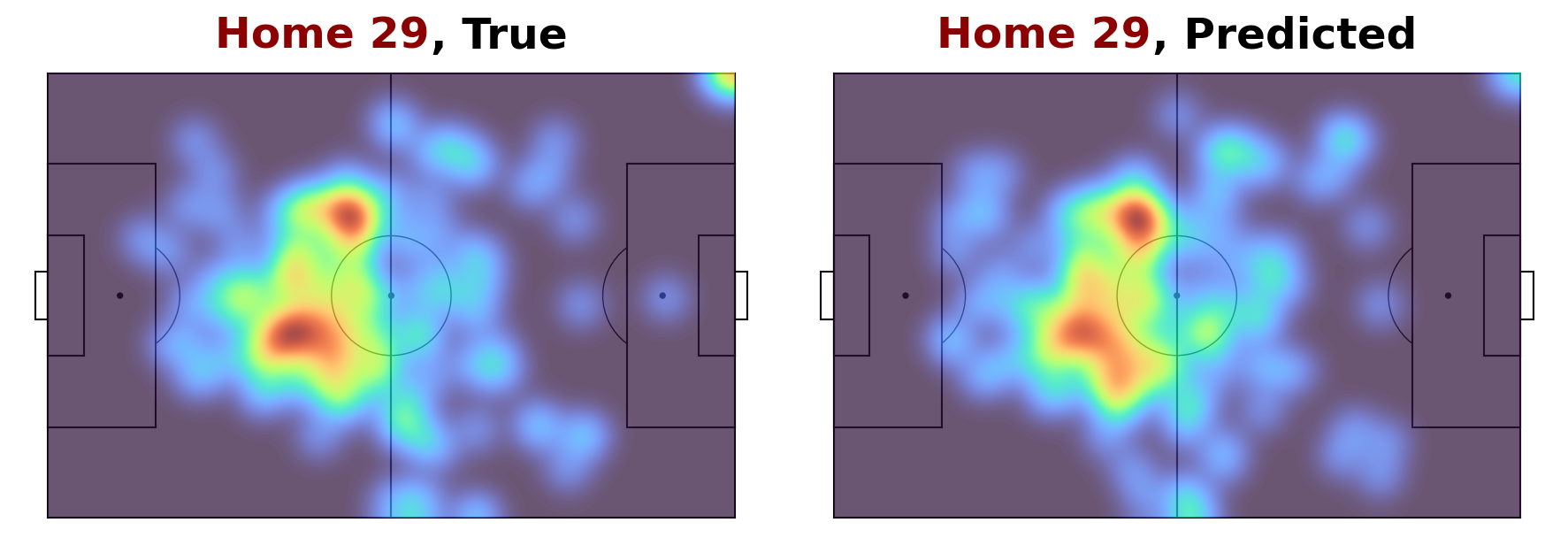}
  \end{minipage}
  \hfill
  \begin{minipage}[t]{0.45\textwidth}
    \includegraphics[width=\linewidth]{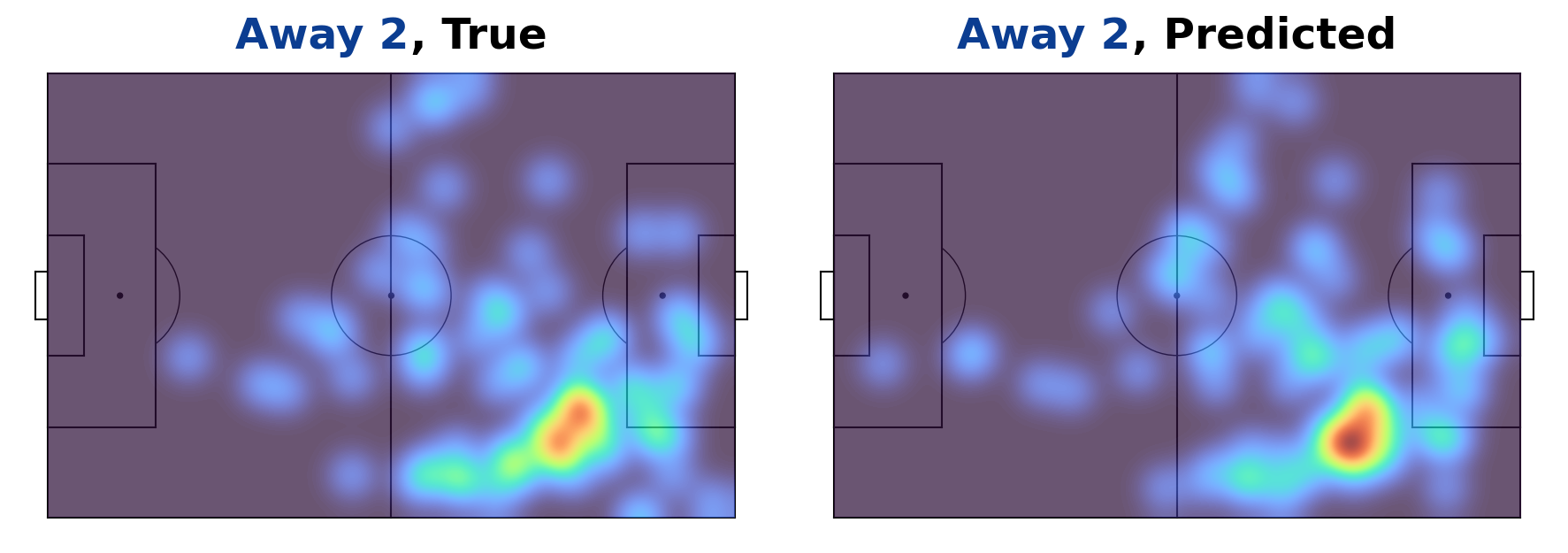}
  \end{minipage}
  \vspace{-1em}
  \caption{Comparison of kernel density estimation (KDE) heatmaps computed from true and detected events, for both teams and the most event-involved player from each team.}
  \vspace{-0.5em}
  \label{fig:heatmaps}
\end{figure}

\subsection{Spatial Distribution of On-Ball Plays}
\label{sec:heatmaps}
Since the spatial distribution of on-ball plays reflects team tactics and individual playing styles, we examine whether PathCRF can accurately reproduce these positional tendencies from detected events. As shown in Fig.~\ref{fig:heatmaps}, the predicted heatmaps for both teams and individual players closely align with the true heatmaps, implying that PathCRF preserves spatial patterns of events with high fidelity. This suggests that a wide range of downstream analyses for characterizing the playing styles at both the team \cite{LuceyOCRM13,DecroosHD18} and player levels \cite{DecroosD19,DecroosVD20} can be directly applied to our detected events, enabling scalable event-based soccer analytics without manual tagging.

\subsection{Team Possession Statistics}
\label{sec:poss_stats}
Team-level possession share is one of the most widely used indicators in both broadcasting and analytics, as it provides a simple yet informative summary of which team controlled the flow of the match. We therefore examine whether team possession can be accurately recovered using only the events detected by PathCRF.

\begin{figure}[t]
    \centering
    \includegraphics[width=.45\textwidth]{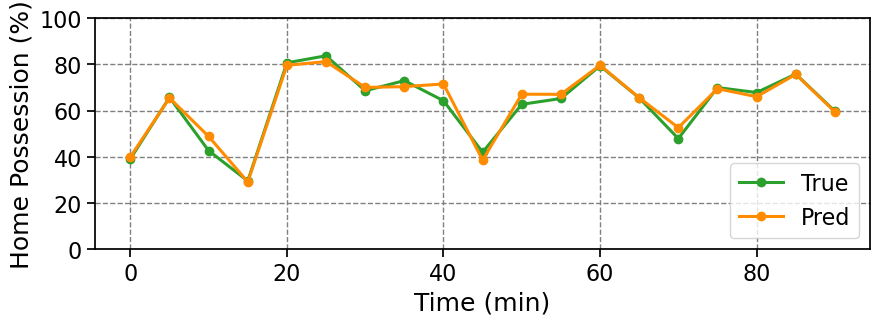}
    \vspace{-0.8em}
    \caption{5-minute timeline of home team's possession shares computed from true and detected events, respectively.}
    \vspace{-0.5em}
    \label{fig:poss_timeline}
\end{figure}

On the test match, the ground-truth possession share of the home team is 62.13\%, while our prediction yields 62.64\%, showing that PathCRF can approximate overall possession statistics with near-perfect accuracy. Interestingly, although the frame-level team possession classification accuracy is 92.40\%, misclassifications occur for both teams and tend to cancel out over time, resulting in an even more accurate estimate of the cumulative possession share. This is also reflected in the home team's possession timeline in Fig.~\ref{fig:poss_timeline}, where the predicted values closely track the ground truth even at a fine granularity of 5-minute intervals.

These results indicate that match dominance can be reliably analyzed using only player tracking data and PathCRF. Moreover, we expect that accurate team possession segmentation enables richer contextual analyses without manual annotation, such as separating offensive and defensive phases when calculating physical performance metrics \cite{BradleyA18,JuDHELB22,KimJYK24} or estimating team formations \cite{KimKCYK22,BauerAS23}.

\subsection{Pass Networks}
\label{sec:passmap}
Pass networks are a widely used representation for analyzing teams' collective passing structures, and have been extensively studied in soccer analytics literature \cite{DuchWN10,LopezT13,CintiaRP15,ClementeMKWM15}. To examine whether PathCRF enables such analyses without manual pass annotation, we construct pass networks from the detected kick events and compare them against those derived from ground-truth event data.

\begin{table}[t!]
\centering
\caption{Similarity between ground-truth and predicted pass networks at node degree, edge weight, and graph levels.}
\vspace{-0.7em}
\small
\setlength{\tabcolsep}{3pt}
\begin{tabular}{l|cc|cc|cc}
\toprule
& \multicolumn{2}{c|}{\textbf{Node degree}}
& \multicolumn{2}{c|}{\textbf{Edge weight}}
& \multicolumn{2}{c}{\textbf{Graph structure}} \\
\textbf{Team}
& \textbf{True mean} & \textbf{MAE}
& \textbf{True mean} & \textbf{MAE}
& \textbf{JSD} & \textbf{Spectral} \\
\midrule
Home & 45.19 & 2.64 & 5.18 & 1.08 & 0.0228 & 0.0270 \\
Away & 24.27 & 2.73 & 3.26 & 1.05 & 0.0552 & 0.0138 \\
\bottomrule
\end{tabular}
\vspace{-0.5em}
\label{tab:passmap}
\end{table}

\begin{figure}[t!]
  \centering
  \begin{minipage}[t]{0.45\textwidth}
    \includegraphics[width=\linewidth]{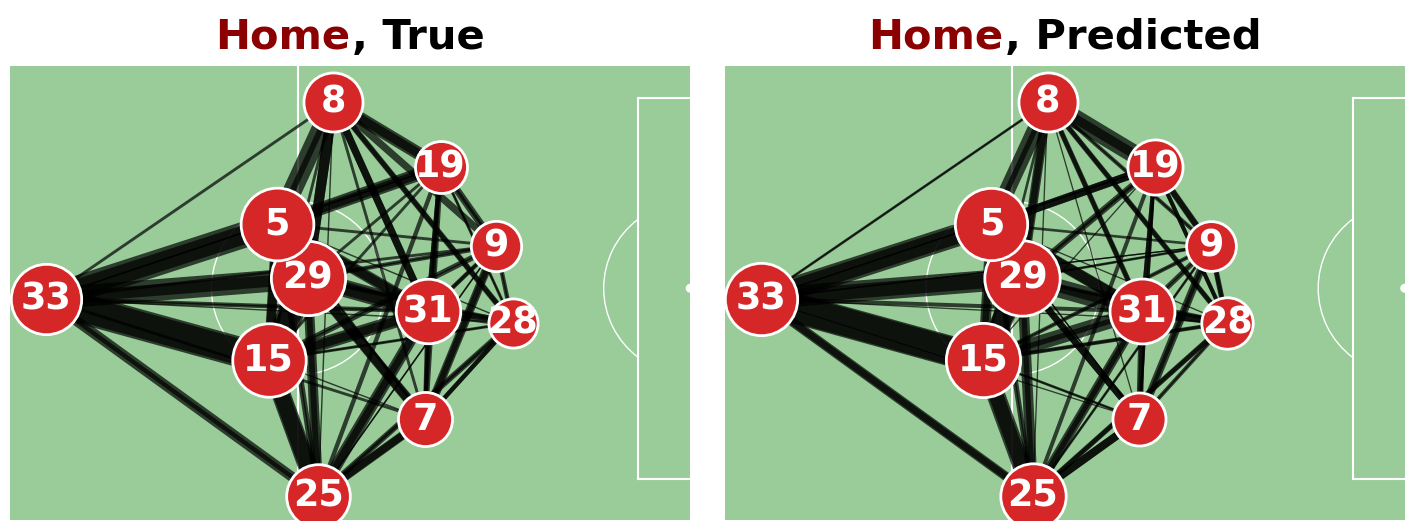}
  \end{minipage}
  \hfill
  \begin{minipage}[t]{0.45\textwidth}
    \includegraphics[width=\linewidth]{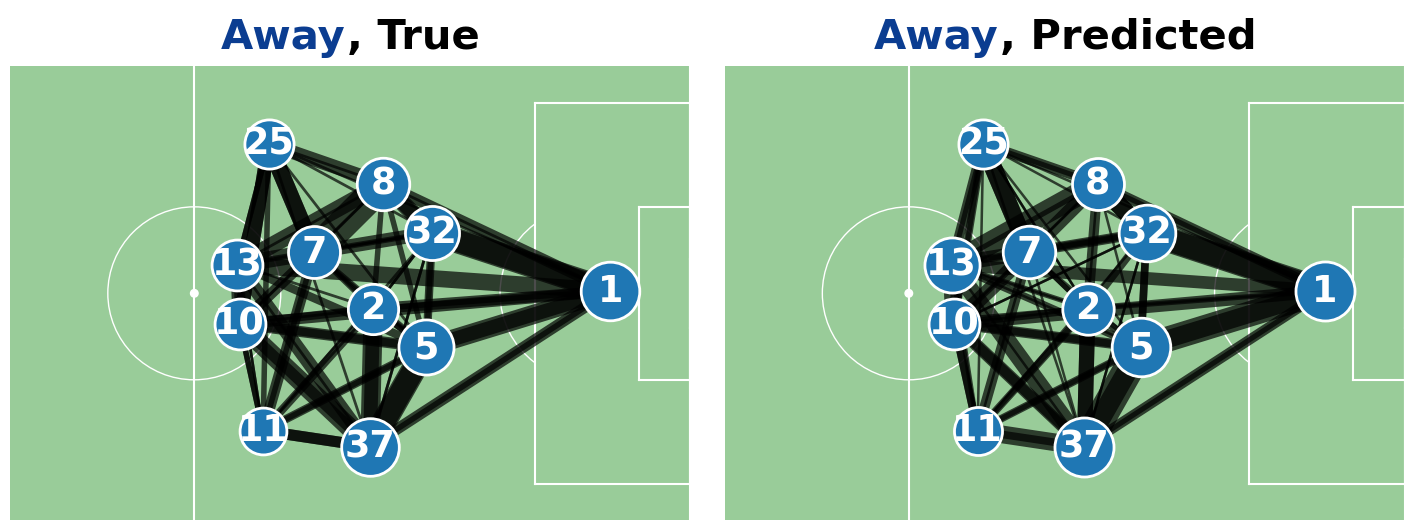}
  \end{minipage}
  \vspace{-0.5em}
  \caption{Visual comparison of pass networks constructed from ground-truth and detected events.}
  \vspace{-0.5em}
  \label{fig:passmap}
\end{figure}

Fig.~\ref{fig:passmap} visualizes the resulting networks for both teams. Each node is positioned at the player’s average location and its size is proportional to the number of successful outgoing passes, where each pair of substituted players are merged as a single node. Edge thickness represents the number of completed passes between the pair of nodes. As shown, the predicted pass networks closely match the ground-truth networks in terms of both the overall passing structure and the key hubs responsible for ball circulation, indicating that PathCRF reliably captures team-level ball flow patterns.

To quantitatively evaluate the similarity between the true and predicted pass networks, we measure how similar they are at the node, edge, and graph levels. Specifically, we compare (i) node degrees and (ii) edge weights using mean absolute error (MAE), and additionally measure (iii) graph-level structural similarity using Jensen-Shannon divergence (JSD) \cite{Lin91} and spectral distance \cite{Chung97}. As reported in Table \ref{tab:passmap}, the MAE values for both node degree and edge weight are small relative to the true mean degrees and weights, respectively, indicating that PathCRF accurately recovers the key passing frequencies. Moreover, both JSD and spectral distance are close to zero, implying that the predicted networks preserve the global structure of team passing patterns. These results demonstrate that diverse downstream analyses, such as quantifying player contributions
\cite{DuchWN10}, team strategies \cite{LopezT13}, and team performance \cite{CintiaRP15,ClementeMKWM15}, can be reliably conducted using detected events.

\subsection{Semi-Automated Event Data Collection}
\label{sec:semi_annot}
Although the events detected by PathCRF can closely approximate various downstream metrics, some fine-grained scene analyses may still require near-perfect event-level accuracy. In such cases, instead of annotating every event from scratch, we propose a human-in-the-loop data collection pipeline where annotators verify and correct model-generated events, substantially reducing the labor cost.

\begin{figure}[t!]
    \centering
    \includegraphics[width=.35\textwidth]{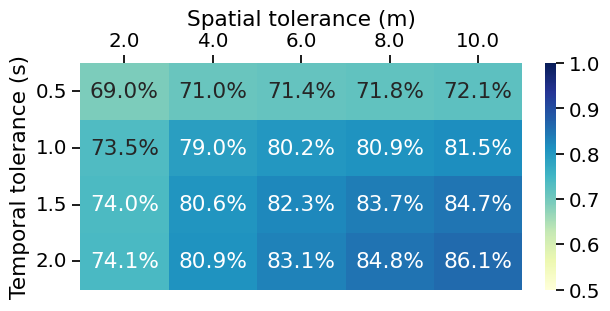}
    \vspace{-0.5em}
    \caption{Event recall under varying tolerance thresholds.}
    \vspace{-1em}
    \label{fig:event_recall}
\end{figure}

Importantly, even when the predicted player identity is incorrect, PathCRF often accurately localizes the event timing and location. To quantify this, we measure recall under relaxed matching criteria, where a detected event is considered correct if its timestamp and location fall within predefined temporal and spatial tolerance thresholds of the ground-truth event. As shown in Fig.~\ref{fig:event_recall}, PathCRF achieves nearly 80\% recall even under a relatively strict criterion of \SI{1}{s} and \SI{4}{m}, which further increases to 86.1\% when allowing discrepancies up to \SI{2}{s} and \SI{10}{m}. These results suggest that many mismatches are near-misses rather than complete failures, which remain useful in practice as they only require correcting a subset of event attributes rather than annotating the entire event records.

Overall, this demonstrates that PathCRF can serve not only as a tool for fully automated event-based analysis, but also as an effective pre-annotation system that significantly streamlines the collection of high-quality soccer event data.

% These results have important implications for practical data collection. Originally, manual event annotation requires recording multiple attributes per event and typically demands several annotators to simultaneously monitor the match. In contrast, our framework enables a human-in-the-loop pipeline where a single annotator can review model-generated events and correct only a small portion (around 20\%) of event attributes, substantially reducing the labor cost of constructing high-quality soccer event datasets.

% As future work, we plan to further reduce the amount of human verification by incorporating additional visual cues from easily accessible broadcast videos into the model.

\section{Conclusion}
\label{sec:conclusion}
This paper proposes PathCRF, a framework for detecting on-ball soccer events using only player tracking data. By formulating event detection as a sequential edge selection problem over a dynamic graph and enforcing hard constraints on edge-to-edge transitions via a dynamic masked CRF, the framework produces logically consistent possession paths and achieves strong event detection performance. Although PathCRF is proposed mainly to replace expensive ball tracking and labor-intensive manual annotation in soccer, its formulation is domain-agnostic. That is, by modifying only the transition constraints defined in Section~\ref{sec:masking}, it can extend to other multi-agent settings such as autonomous driving (detecting vehicle interactions), multi-robot coordination (tracking object handovers), and surveillance (inferring the transfer path of suspicious items).
% Moreover, we show that the detected events can reliably support various downstream analyses, highlighting PathCRF's practical value as a scalable alternative to expensive ball tracking systems and labor-intensive manual event annotation. Beyond soccer, our formulation is domain-agnostic in that it can extend to other multi-agent settings such as autonomous driving, multi-robot coordination, and surveillance, by replacing only the transition constraint set defined in Section~\ref{sec:masking}.

As future work, we aim to further improve detection accuracy by integrating partial human guidance for ambiguous events or visual features from easily accessible broadcast videos into the model. Leveraging such multi-modal information could not only enhance robustness in challenging scenarios but also enable finer-grained categorization of kick events into more detailed action types, such as passes, crosses, shots, and clearances.

\section*{Acknowledgments}
This work was supported by the National Research Foundation of Korea (NRF) grant funded by the Korea government (MSIT) (RS-2024-00335098, RS-2024-00406985), and by the National Research Foundation of Korea (NRF) funded by Ministry of Science and ICT (RS-2022-NR068758).

\section*{GenAI Usage Disclosure}
We used generative AI tools in a limited and supportive manner during the preparation of this manuscript. Specifically, we used ChatGPT to improve the clarity and readability of the writing through phrasing and grammar refinement. In addition, we used Codex and Claude Code to support data preprocessing, visualization, and debugging. All core research contributions, including the main ideas, methodology design, experiments, and analytical insights, were developed and validated entirely by the authors.

\bibliographystyle{ACM-Reference-Format}
\balance
\bibliography{main}

\appendix
\section{Forward and Viterbi Algorithms}
In this section, we elaborate on the forward and Viterbi algorithms for Conditional Random Fields (CRFs) \cite{LaffertyMP01}, which are adapted to handle hard structural constraints in the proposed PathCRF.

\subsection{Forward Algorithm for Training}
\label{sec:forward}
Recall that the total score of an edge sequence $\mathbf{e}=(e_1, \ldots, e_T)$ is defined in Eq. \ref{eq:score} as:
\begin{equation}
    S(\mathbf{e}) = f_1(e_1) + \sum_{t=2}^T \left( f_t(e_t) + \psi_t(e_{t-1}, e_t) \right).
\end{equation}
where the emission scores $f_t(e_t)$ and the transition scores $\psi(e_{t-1}, e_t)$ are derived from the edge embeddings. Using this score, Eq. \ref{eq:crf_loss} defines the CRF loss as the negative log-likelihood of the ground-truth edge sequence $\mathbf{e}^* = (e^*_1, \ldots, e^*_T)$ given the input trajectories $\mathbf{X}$:
\begin{equation}
\mathcal{L}_{\text{CRF}} = -\log p(\mathbf{e}^* | \mathbf{X}) = \log Z - S(\mathbf{e}^*).
\end{equation}

To minimize $\mathcal{L}_{\text{CRF}}$, we must compute the partition function $Z = \sum_{\mathbf{e'}} \exp(S(\mathbf{e'}))$. Since summing over all possible sequences is computationally intractable, ($\mathcal{O}(|\mathcal{E}|^T)$ where $\mathcal{E}$ is the set of edges), we use the forward algorithm to effectively compute $\log Z$ via dynamic programming.

Let $\alpha_t(e)$ be the forward variable, representing the log-sum-exp score of all partial sequences ending with edge $e$ at time $t$:
\begin{equation}
    \alpha_t(e) = \log \sum_{\mathbf{e}_{1:t}, e_t=e} \exp(S(\mathbf{e}_{1:t})).
\end{equation}
The values of $\alpha_t(e)$ for $e \in \mathcal{E}$ and $t = 1, \ldots, T$ are obtained via the following recursion, and $\{\alpha_T(e): e \in \mathcal{E}\}$ at the last time step are used to calculate $\log Z$.

\vspace{1mm}
\noindent \textbf{Initialization ($t=1$).} At the first time step, $\alpha_1(e)$ is simply the emit score of $e$:
\begin{equation}
    \alpha_1(e) = f_1(e), \quad \forall e \in \mathcal{E}.
\end{equation}

\vspace{1mm}
\noindent \textbf{Recursion ($t=2, \ldots, T$).} We compute $\alpha_t(e)$ by aggregating scores from the previous time step. In particular, to enforce the hard constraints defined in Section \ref{sec:masking}, we iterate only over the set of allowed previous edges $\mathcal{P}(e) = \{e'\in\mathcal{E}: (e',e)\in \mathcal{A}\}$:
\begin{equation}
    \alpha_t(e) = f_t(e) + \log \sum_{e'\in\mathcal{P}(e)} \exp (\alpha_{t-1}(e') + \psi_t(e',e)).
\end{equation}
Although the summation is in principle taken over all edges $e' \in \mathcal{E}$, $\psi_t(e',e) = -10^4$ for illegal transitions $(e',e) \notin \mathcal{A}$, making their terms negligible, i.e., $\exp (\alpha_{t-1}(e') + \psi_t(e',e)) \approx 0$.

\vspace{1mm}
\noindent \textbf{Termination.} Finally, $\log Z$ is obtained as the log-sum-exp of the forward variables at the last time step $T$:
\begin{equation}
    \log Z = \log \sum_{e\in\mathcal{E}} \exp(\alpha_T(e)).
\end{equation}
This reduces the complexity to $\mathcal{O}(T\cdot|\mathcal{A}|)$, which is even smaller than the forward algorithm of standard dense CRFs with $\mathcal{O}(T\cdot|\mathcal{E}|^2)$ due to the sparsity of $\mathcal{A}$.

\subsection{Viterbi Algorithm for Inference}
\label{sec:viterbi}
During inference, our goal is to find the most probable edge sequence $\hat{\mathbf{e}} = \argmax_{\mathbf{e}} S(\mathbf{e})$, which is equivalent to finding the sequence that maximizes the score $S(\mathbf{e})$.

Let $\delta_t(e)$ be the max-score variable, representing the highest score among all partial sequences ending with edge $e$ at time $t$:
\begin{equation}
    \delta_t(e) = \max_{\mathbf{e}_{1:t}, e_t=e} S(\mathbf{e}_{1:t}).
\end{equation}
The values of $\delta_t(e)$ for $e \in \mathcal{E}$ and $t = 1, \ldots, T$ are obtained via the following recursion, and the optimal sequence is extracted by backtracking the optimal predecessor from $t=T$ to $t=1$.

\vspace{1mm}
\noindent \textbf{Initialization ($t=1$).} At the first time step, $\delta_1(e)$ is simply the emit score of $e$:
\begin{equation}
    \delta_1(e) = f_1(e), \quad \forall e \in \mathcal{E}.
\end{equation}

\vspace{1mm}
\noindent \textbf{Recursion ($t=2, \ldots, T$).} We compute the maximum score reaching each edge $e$ at $t$ while recording the optimal predecessor of $e$ in a backpointer table $\text{ptr}_t(e)$:
\begin{align}
    \delta_t(e) = f_t(e) + \max_{e'\in\mathcal{P}(e)} (\delta_{t-1}(e') + \psi_t(e',e)) \\
    \text{ptr}_t(e) = \argmax_{e'\in\mathcal{P}(e)} (\delta_{t-1}(e') + \psi_t(e',e))
\end{align}
By restricting the search space to $\mathcal{P}(e)$, we ensure the selected path never contains illegal transitions.

\vspace{1mm}
\noindent \textbf{Termination and backtracking.} We first identify the edge with the highest score at the final time step:
\begin{equation}
    \hat{e}_T = \argmax_{e\in\mathcal{E}} \delta_T(e).
\end{equation}
Then, we reconstruct the optimal sequence $\hat{\mathbf{e}} = \argmax_{\mathbf{e}} S(\mathbf{e})$ by backtracking via the pointers:
\begin{equation}
    \hat{e}_{t-1} = \text{ptr}_t(\hat{e}_t), \quad t = T, \ldots, 2.
\end{equation}
The resulting sequence $\hat{\mathbf{e}}$ is the globally optimal possession path under the learned constraints.

\section{Cross-Dataset and Cross-Match Robustness}
\label{app:robustness}
To evaluate our framework beyond the single-match test data, we additionally acquire a proprietary dataset of \SI{10}{Hz} GPS tracking data and event annotations collected from 15 matches of South Korean K League 1 (10:2:3 train/validation/test split). For the Sportec dataset, we further perform leave-one-match-out cross-validation (LOMOCV), where each of the 7 matches serves as the test set in turn, with 5 matches for training and 1 for validation.

As shown in Table~\ref{tab:robustness} (rows 1 and 2), PathCRF performs well on both providers when trained and evaluated within the same dataset. In addition, the larger K League training set yields higher performance (row 1), suggesting that performance scales with more training data, while the small standard deviations of the Sportec LOMOCV (row 2) confirm robustness across different test matches. In contrast, when a model trained on one dataset is evaluated on the other (rows 3 and 4), performance drops due to cross-provider distribution shift. That is, the two providers differ in collection modality (wearable GPS vs. camera-based tracking) and annotation conventions, producing distinct noise patterns and ground-truth labels. Nonetheless, this is not a critical limitation in practice, as the intended users of PathCRF are data providers who would train on their own data to reduce future annotation costs, for which cross-provider generalization is not required.

\begin{table}[t]
\centering
\footnotesize
\setlength{\tabcolsep}{2pt}
\caption{Cross-dataset and LOMOCV performance, reported as the mean $\pm$ standard deviation across test matches.}
\label{tab:robustness}
\vspace{-0.8em}
\begin{tabular}{ll|cccc}
\toprule
\textbf{Train.} & \textbf{Test} & \textbf{Edge acc.} & \textbf{Event prec.} & \textbf{Event recall} & \textbf{Event F1} \\
\midrule
K League & K League & $68.20 \pm 2.64\%$ & $84.34 \pm 2.42\%$ & $75.79 \pm 2.76\%$ & $79.83 \pm 2.60\%$ \\
Sportec & Sportec & $66.75 \pm 2.76\%$ & $72.05 \pm 2.37\%$ & $75.87 \pm 2.33\%$ & $73.88 \pm 1.90\%$ \\
K League & Sportec & $52.71 \pm 9.31\%$ & $71.93 \pm 6.43\%$ & $61.52 \pm 8.41\%$ & $66.25 \pm 7.65\%$ \\
Sportec & K League & $41.65 \pm 2.34\%$ & $54.44 \pm 3.68\%$ & $44.84 \pm 3.28\%$ & $49.16 \pm 3.37\%$ \\
\bottomrule
\end{tabular}
\end{table}

\section{Ablation Study on Loss Terms}
\label{sec:ablation_loss}
To examine the impact of the auxiliary loss terms introduced in Eq.~\ref{eq:loss}, we conduct an ablation study by selectively removing $\mathcal{L}_{\text{coarse}}$ and $\mathcal{L}_{\text{emit}}$ from the training objective, by setting $\lambda_1 = 0$ or $\lambda_2 = 0$. We perform this experiment on both Static and Dynamic Masked CRFs, and report the results in Table \ref{tab:ablation_loss}, where the best-performing configuration for each CRF variant is highlighted in \textbf{bold}.

\begin{table}[bt]
\centering
\footnotesize
\caption{Loss function ablation study on the test data.}
\label{tab:ablation_loss}
\vspace{-0.8em}
\setlength{\tabcolsep}{2pt}
\begin{tabular}{l|ccc|cccc}
\toprule
\textbf{CRF module}
& $\mathcal{L}_{\text{CRF}}$
& $\mathcal{L}_{\text{coarse}}$
& $\mathcal{L}_{\text{emit}}$
& \makecell[c]{\textbf{Edge} \\ \textbf{acc.}}
& \makecell[c]{\textbf{Event} \\ \textbf{prec.}}
& \makecell[c]{\textbf{Event} \\ \textbf{recall}}
& \makecell[c]{\textbf{Event} \\ \textbf{F1}} \\
\midrule
\multirow{4}{*}{Static MCRF}
& \cmark & \cmark & \cmark
& \textbf{69.54\%} & 68.57\% & \textbf{79.96\%} & \textbf{73.83\%} \\
& \cmark & \xmark & \cmark
& 66.96\% & 62.66\% & 79.36\% & 70.02\% \\
& \cmark & \cmark & \xmark
& 66.09\% & \textbf{69.26\%} & 75.91\% & 72.43\% \\
& \cmark & \xmark & \xmark
& 65.41\% & 66.19\% & 73.13\% & 69.49\% \\
\midrule
\multirow{4}{*}{Dynamic MCRF}
& \cmark & \cmark & \cmark
& \textbf{69.64\%} & \textbf{73.18\%} & \textbf{78.37\%} & \textbf{75.69\%} \\
& \cmark & \xmark & \cmark
& 67.52\% & 70.02\% & 75.75\% & 72.77\% \\
& \cmark & \cmark & \xmark
& 65.77\% & 67.48\% & 74.33\% & 70.74\% \\
& \cmark & \xmark & \xmark
& 65.64\% & 71.32\% & 68.05\% & 69.65\% \\
\bottomrule
\end{tabular}
\end{table}

In Section \ref{sec:results}, the baseline comparison reported in Table~\ref{tab:results} shows that edge-level accuracies remain largely similar across different structuring methods and performance differences mainly emerge at the event level. In contrast, Table~\ref{tab:ablation_loss} demonstrates that removing auxiliary losses significantly degrades the accuracy from the edge level, which leads to weaker event detection performance. This indicates that the auxiliary terms play a crucial role in guiding the backbone toward learning discriminative possession representations, thereby stabilizing CRF training and improving both edge selection and downstream event detection. Overall, these results confirm that auxiliary supervision is essential for achieving strong performance in our framework.

\section{Related Work}
\subsection{Trajectory Modeling in Multi-Agent Sports}
Modeling trajectories in fluid multi-agent sports such as soccer and basketball is challenging due to the dynamic interactions among many agents. Moreover, since players have no semantic ordering, effective models must satisfy permutation-equivariance with respect to the input players. Early studies~\cite{FelsenLG18,LuceyBCMMS13,ZhanZYSL19} impose a fixed player ordering based on detected tactical roles~\cite{BialkowskiLCYSM14}, but they are fundamentally limited because players cannot be consistently aligned into a single ordering across diverse formations. Subsequent work adopted Graph Neural Networks (GNNs)~\cite{YehSHM19,OmidshafieiHGWRT22,XuBCCF23} and their graph-attention variants~\cite{FassmeyerFB22,MontiBCC20,LiYTC20,SunKZLKWH22} to naturally capture interactions while preserving permutation-equivariance. More recently, Transformer~\cite{VaswaniSPUJGKP17} and Set Attention~\cite{LeeLKKCT19} have become popular alternatives, as they enable fast tensorized computation leveraging the fixed maximum number of agents~\cite{AlcornN21A,AlcornN21B,CapelleraFRAM24,ChoiKLJKYK25,HughesHWGFSL25,JoHCLBK25,KimCKYK23,PeralCFRA26}, whereas GNNs rely on computationally inefficient graph batching.

% Subsequent work increasingly adopted Graph Neural Networks (GNNs)~\cite{EverettBMENR23,YehSHM19,OmidshafieiHGWRT22,XuBCCF23}, which naturally capture player interactions while preserving permutation-equivariance. Their variants based on graph attention~\cite{DingH20,FassmeyerFB22,MontiBCC20,LiYTC20,SunKZLKWH22} further improved expressiveness by adaptively weighting the influence of neighboring agents depending on the context.

From a task perspective, sports trajectory understanding can be categorized into trajectory forecasting, trajectory imputation, ball trajectory inference, and state classification~\cite{CapelleraFRAM24}. Among them, trajectory forecasting has inherent uncertainty, where multiple plausible futures exist for the same input. Hence, it is often approached with generative models such as Variational Recurrent Neural Networks~\cite{YehSHM19,ZhanZYSL19,LiYTC20,MontiBCC20,FassmeyerFB22,SunKZLKWH22} and diffusion models~\cite{CapelleraRFA25,CapelleraFRAA26}. In contrast, ball trajectory inference~\cite{KimCKYK23,CapelleraFRAM24,PeralCFRA26} and state classification~\cite{BauerAS23,CapelleraFRAM24,PeralCRFMA25,PeralCFRA26} are typically treated as deterministic problems, as ball dynamics and on-ball events are largely determined once player movements are fully observed. As a possession state classification framework, our work follows this deterministic perspective but differs by explicitly enforcing logical consistency between consecutive states through a CRF-based architecture.

\subsection{Structured Sequence Inference}
Structured sequence labeling is the task of assigning a label to each token of an input sequence while satisfying predefined structural constraints~\cite{HeWWFMJ20}. A canonical example is the Beginning-Inside-Outside (BIO) tagging scheme used in Named Entity Recognition (NER), where an `inside' (I) tag can only follow either a `beginning' (B) tag or another `I' tag of the same entity. To model such dependencies, several approaches~\cite{HuangXY15,MaH16,LampleBSKD16,ZhengJRVSDHT15} employ CRFs~\cite{LaffertyMP01} on top of LSTMs~\cite{HochreiterS97} to globally optimize the label sequence rather than making per-token local predictions. To better capture the time-varying context, some studies~\cite{HongL20,ThaiRMVM18} dynamically adapt transition probabilities instead of using a static transition matrix. Furthermore, others~\cite{LesterPHCB20,WeiQHS21,PapayKP22} explicitly mask illegal transitions in the CRF architecture, thereby guaranteeing structurally valid outputs.

% To explicitly model such dependencies, Huang et al.~\cite{HuangXY15} employs a CRF~\cite{LaffertyMP01} architecture on top of Bi-LSTMs~\cite{HochreiterS97} to globally optimize the label sequence rather than per-token local predictions. Subsequent work~\cite{MaH16,LampleBSKD16,ZhengJRVSDHT15} further enhanced this framework by incorporating character-level representations via CNNs or RNNs to capture morphological features without manual feature engineering. Some studies~\cite{HongL20,ThaiRMVM18} have explored dynamically adapting transition probabilities based on the input context instead of static transition matrix to better capture their time-varying dependencies. Furthermore, several approaches~\cite{LesterPHCB20,WeiQHS21,PapayKP22} explicitly mask illegal transitions in the CRF architecture, thereby guaranteeing structurally valid outputs by construction.

More recently, constrained decoding has gained attention in the context of large language models (LLMs), where enforcing structural constraints during training is difficult due to the autoregressive formulation and the extremely large output spaces that make traditional CRFs intractable \cite{SunLWHLD19}. Instead, many approaches impose hard constraints directly at inference time \cite{HokampL17,LuWWJKKBQYZSC22,GengJP023}, ensuring legal outputs without modifying the underlying model.

% Inspired by these advancements in NLP, we apply the CRF framework to sports event detection, a domain where sequential constraints are equally critical but less explored. While Wang et al.~\cite{WangASF14} previously utilized a CRF for ball tracking in sports video, our work addresses a distinct challenge of detecting events solely from player trajectory data without visual cues. We distinguish our framework by integrating a socio-temporal deep learning backbone with a dynamic CRF, enabling robust and logically consistent event detection even in the absence of ball tracking data.

\end{document}